\documentclass{article}
\usepackage{rotating}
\usepackage{multirow}
\usepackage{subfigure}
\usepackage{subcaption}
\usepackage[dvipsnames]{xcolor}

\usepackage[numbers,sort&compress]{natbib}

\usepackage[main, final]{neurips_2025}

\usepackage[utf8]{inputenc} 
\usepackage[T1]{fontenc}    
\usepackage{hyperref}       
\usepackage{url}            
\usepackage{booktabs}       
\usepackage{amsfonts}       
\usepackage{nicefrac}       
\usepackage{microtype}      
\usepackage{xcolor}         
\usepackage{graphicx}
\usepackage{enumitem}
\usepackage{amsmath} 
\usepackage{mdframed}
\usepackage[most]{tcolorbox}
\definecolor{mydarkgreen}{RGB}{0,100,0}
\tcbuselibrary{breakable}
\newtcolorbox{mycustombox}[1]{
  enhanced,
  colback=black!5!white,
  colframe=black!75!white,
  boxrule=0.4pt,
  coltitle=white,
  title=#1,
  titlerule=0.4pt,
  fontupper=\small,
  fonttitle=\small,
  before upper={\par\smallskipamount},
  breakable,
  left=3mm,
  right=3mm,
  top=2mm,
  bottom=2mm,
  boxsep=1mm,
}

\usepackage{subcaption}

\title{Baichuan-M2: Scaling Medical Capability with Large Verifier System}

\author{%
  Baichuan-M2 Team
}

\begin{document}

\maketitle

\begin{abstract}
As large language models~(LLMs) advance in conversational and reasoning capabilities, their practical application in healthcare has become a critical research focus. However, there is a notable gap between the performance of medical LLMs on static benchmarks such as USMLE and their utility in real-world clinical decision-making. This discrepancy arises because traditional exams fail to capture the dynamic, interactive nature of medical consultations. To address this challenge, we introduce a novel dynamic verification framework that moves beyond static answer verifier, establishing a large-scale, high-fidelity interactive reinforcement learning system. Our framework comprises two key components: a \textit{Patient Simulator} that creates realistic clinical environments using de-identified medical records, and a \textit{Clinical Rubrics Generator} that dynamically produces multi-dimensional evaluation metrics. Building on this foundation, we develop Baichuan-M2, a 32B-parameter medical augmented reasoning model trained through a multi-stage reinforcement learning strategy with an improved Group Relative Policy Optimization (GRPO) algorithm. Evaluated on HealthBench, Baichuan-M2 outperforms all other open-source models and most advanced closed-source counterparts, achieving a score above 32 on the challenging HealthBench Hard benchmark—previously exceeded only by GPT-5. Our work demonstrates that robust dynamic verifier system is essential for aligning LLM capabilities with practical clinical applications, establishing a new Pareto front in the performance-parameter trade-off for medical AI deployment.
\end{abstract}

\section{Introduction}

As the conversational and reasoning capabilities of large language models (LLMs) continue to advance, there is increasing interest in their practical application in specific domains. The healthcare sector, in particular, has become a key area of research, attracting significant investment from both global tech giants and innovative startups~\cite{GoyalRRYZCNW24,CascellaMBB23,YangXYRZISGW24}. Among the various approaches aimed at enhancing the capabilities of LLMs in healthcare, reinforcement learning with verifiable rewards (RLVR) has garnered considerable attention~\cite{liu2025beyond,chen2024huatuogpt}. This technique has already demonstrated impressive results in areas such as mathematics~\cite{guo2025deepseek,jaech2024openai,shao2024deepseekmath}, code~\cite{anthropic2025introducingclaude4}, agents~\cite{team2025kimi,zeng2025glm}, and multimodality~\cite{hong2025glm,lasa2025lingshu}. These achievements highlight its potential to significantly enhance model reasoning, making its application in healthcare a highly promising research direction.

The core of RLVR lies in the development of a robust evaluation system. Its success in fields like mathematics and coding can be attributed to the availability of precise and reliable evaluation metrics. However, when it comes to assessing LLMs in the medical domain, a significant gap exists between current evaluation methods and real-world applications. Models that perform well on medical professional exams, such as the USMLE~\cite{usmle_scoring_policy_2024}, often underperform in practical clinical decision-making. This discrepancy arises because traditional static benchmarks fail to capture the dynamic and complex nature of clinical practice. Real-world medical consultations frequently involve incomplete information, multiple rounds of diagnostic exploration, and nuanced communication skills, all of which are not adequately measured by conventional exams.

To address these challenges, we transitioned our focus from static answer verifiers to the development of a large-scale, high-fidelity interactive reinforcement learning verifier system. This system transcends conventional answer verifier by simulating real-world clinical scenarios, allowing the model to learn and adapt through simulated ``practice'' in a virtual clinical environment. Building on this foundation, we introduce Baichuan-M2, a medical augmented reasoning model that marks a significant advancement in open-source medical artificial intelligence.

Specifically, our verifier system comprises two key components. The first is a Patient Simulator, which integrates desensitized medical records and doctor-patient conversation records to effectively simulate patients with diverse social backgrounds and personality traits. This provides a highly realistic interactive environment. The second component is a Clinical Rubrics Generator, which can emulate the clinical reasoning of experienced doctors. It dynamically generates quantifiable evaluation rubrics on a large scale, based on multiple dimensions such as diagnostic accuracy, consultation logic, treatment plan rationality, communication empathy, and medical ethics.

Our training process includes mid-training for medical domain adaptation, supervised fine-tuning (SFT) with rejection sampling, and multi-stage reinforcement learning (RL) using an improved Group Relative Policy Optimization (GRPO)~\cite{shao2024deepseekmath} algorithm.
Specifically, we employ a multi-stage reinforcement learning strategy to decompose complex reinforcement learning tasks into a controllable hierarchical structure. This approach enhances various capabilities, including medical knowledge, reasoning, and patient interaction, while maintaining the general capabilities of the model.

We evaluate our model on the challenging HealthBench dataset~\cite{arora2025healthbenchevaluatinglargelanguage}, developed by OpenAI. Despite its relatively small number of parameters (only 32B), Baichuan-M2 outperformed all other open-source models, including gpt-oss-120B, and most advanced closed-source counterparts on HealthBench. It particularly excelled on the HealthBench Hard test, achieving a score exceeding 32, a performance level previously reached by only one other model globally, GPT-5. These experimental results underscore the critical role of a robust validation system in integrating model capabilities with practical applications.

In summary, our contributions can be highlighted as follows:
\begin{itemize}[left=0pt]
\item A dynamic verifier system tailored for clinical scenarios, which addresses the limitations of previous verification methods based on static data. This method employs a patient simulator to create a high-fidelity decision-making environment and uses clinical rubrics to generate quantitative evaluation metrics in real time, thereby enhancing the reliability of the verification process.

\item An advanced training method that successfully implements a multi-stage reinforcement learning strategy in a dynamic interactive environment, featuring targeted improvements to the GRPO algorithm. This enhancement enables the model to move beyond static knowledge memorization and deeply align with the advanced clinical reasoning capabilities of medical experts.

\item An advanced open-source model, Baichuan-M2, which achieves top-tier performance at a remarkably lower deployment cost, setting a new Pareto front in the performance-parameter trade-off. This efficiency makes the deployment of advanced medical AI more feasible in resource-constrained healthcare settings.
\end{itemize}

\section{Verifier System}

\begin{figure}
    \centering
    \includegraphics[width=0.9\linewidth]{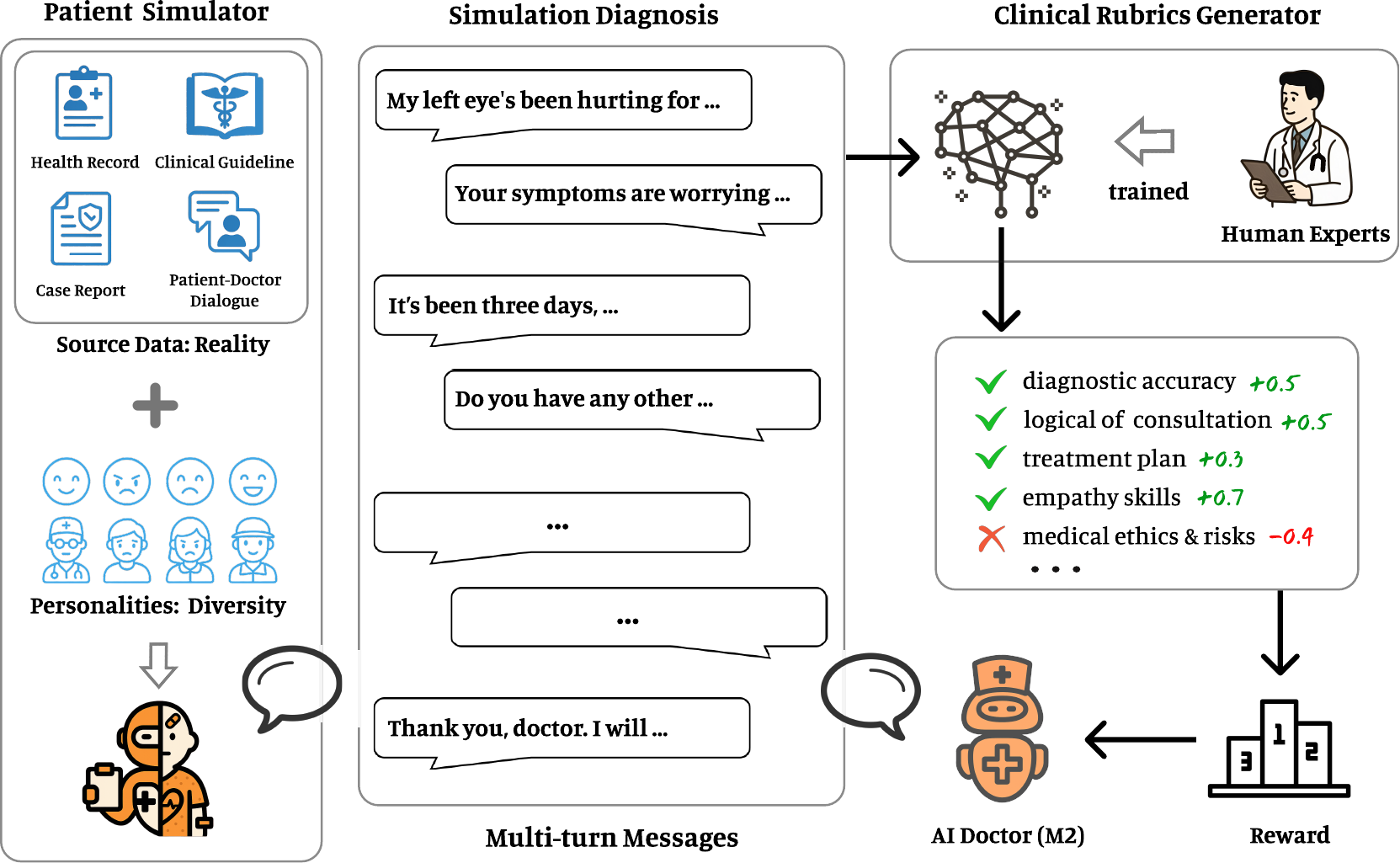}
    \caption{Verifier System Framework}
    \label{fig:framework}
\end{figure}

In recent years, RLVR has achieved remarkable success in complex reasoning domains such as mathematics, coding, and agentic systems. Constructing more verifiable complex problems and environments has become a core driver for continuous breakthroughs in model capabilities. However, when applying this paradigm to the medical field, we discovered significant limitations: static answer verifier built on traditional medical question banks fails to capture the dynamic complexity of real-world diagnostic processes, often leading to limited generalization and suboptimal performance in practical applications. Real clinical practice is a partial observable, multi-turn decision-making process that relies heavily on a physician's dynamic judgment, entailing the integration of clinical experience, communication skills, and ethical considerations.

To address this challenge, in the development of Baichuan-M2, we shifted our focus from building static answer verifiers to creating a large-scale, high-fidelity dynamic interactive reinforcement learning environment. This environment aims to construct a ``virtual clinical world'' where models can ``train and grow''. The system primarily consists of two key modules: a ``patient simulator'' and a ``clinical rubrics generator''. The patient simulator elevates the training environment beyond rigid single-turn QA, generating realistic, stochastic, continuous interaction scenarios. The clinical rubrics generator dynamically produces verification rules for answers, enabling continuous and dynamic quantitative assessment of a model's comprehensive performance across multi-turn interactions as shown as Figure~\ref{fig:framework}.

Through this closed-loop system, we successfully implemented large-scale end-to-end reinforcement learning. The model continuously interacts with ``virtual patients'', iteratively optimizing its diagnostic strategies based on dense feedback from ``expert-level evaluations''. Ultimately, the model's capabilities move beyond recall of static knowledge, achieving deep alignment with the clinical thinking and practical skills of senior physicians.

\subsection{Patient Simulator}
Patient simulators play a critical role in the training and evaluation of AI physicians~\cite{liu2025exploring, porcar2025patient}. These simulators offer a dynamic testing environment that can effectively address the limitations of traditional static testing methods, which often fail to adequately assess the dynamic diagnostic capabilities of LLMs. However, widely used simulators in prior works~\cite{li2025agenthospitalsimulacrumhospital, wang2025surveyllmbasedagentsmedicine} fall short in comprehensively modeling patients' psychological states, social backgrounds, and dynamic interactions. This deficiency reduces these simulators to static databases, thereby limiting their ability to replicate the complexity of real-world clinical encounters. Such encounters often involve information withholding, emotional expressions, and culturally-mediated communication barriers, all of which are crucial for the adaptability of AI physicians in practical settings.

The core challenge in developing high-fidelity patient simulators lies in balancing diversity and consistency. Achieving diversity necessitates an extensive disease knowledge base coupled with multidimensional behavior models to cover broad clinical scenarios. Conversely, ensuring consistency requires preset scripts and behavioral constraints to maintain reproducibility for specific cases.

Building on prior research~\cite{liu2025exploring}, we trained a high-fidelity patient simulator that achieves an optimal diversity-consistency tradeoff, providing a highly realistic interactive environment.

\subsubsection{Patient Scripts}
Patient scripts integrate medical and psychological information to enhance behavioral simulation.

\paragraph{Medical Information.} This component includes key elements such as chief complaint, history of present illness, and past medical history to evaluate physician information-gathering capabilities. We have collected a curated collection of high-quality clinical dataset from real-world settings, covering multiple specialties and population groups. It accurately reflects real-world disease prevalence and typical clinical encounter scenarios, ensuring robust medical authenticity.

\paragraph{Psychological Information.} Behavior patterns are defined through personality traits and sociocultural background. Inspired by the MBTI 16-type model~\cite{myers1962myers}, we mapped distinct behavioral manifestations, For example: extroverts (E) proactively inquire about treatments, while introverts (I) passively accept information; feeling types (F) exhibit greater sensitivity to communication style than thinking types (T), subsequently affecting treatment compliance. Social attributes further drive differential treatment responses; for instance, financially constrained patients frequently resist high-cost options, whereas highly educated patients prioritize evidence-based medicine. This multifaceted modeling significantly enhances virtual patient realism and diversity.

\subsubsection{Modules and Inner-interaction}
During the implementation of the patient simulator, we observed that larger models exhibited higher persona fidelity but incurred prohibitive computational costs, limiting their integration into reinforcement learning training loops. Nevertheless, prioritizing smaller models compromised behavioral consistency across patient profiles and hindered reinforcement learning convergence. 
Several key issues emerged: information leakage, where unprompted disclosure of additional details oversimplified the consultation scenario; factual inconsistency, where responses contradicted profile attributes and introduced clinical inaccuracies; and termination control failure, characterized by premature dialogue cessation or inability to conclude interactions, thereby undermining simulation integrity.

\begin{figure}
    \centering
    \includegraphics[width=0.7\linewidth]{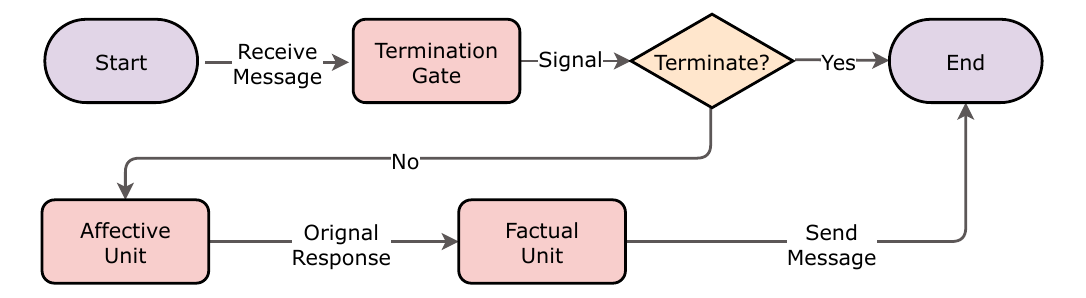}
    \caption{An illustration of Patient Simulator. The system is composed of three primary modules: the Termination Gate, the Affective Unit, and the Factual Unit. The Affective Unit was trained using synthetic data to simulate patients with a wide range of personalities and sociocultural backgrounds. Both the Affective Unit and the Factual Unit were implemented via LLMs. These units employ a non-thinking model to quickly determine termination conditions and verify factual information.}
    \label{fig:sp_structure}
\end{figure}

To address these challenges, we propose a three-component architecture (Figure~\ref{fig:sp_structure}) comprising: a Termination Gate that determines conversation conclusion based on predefined triggers (e.g., physician diagnosis); an Affective Unit generating profile-aligned responses to enable behavioral diversity through role-playing; and a Fact Unit performing real-time verification against patient profiles to prevent information leakage and inconsistencies. Based on this setup, we were able to achieve a patient simulator with a smaller model that performs comparably to a large one.

\subsubsection{Performance of Patient Simulator}
We propose a dual-dimensional evaluation framework integrating granular turn-based analysis with holistic session-level fidelity metrics. At the single-turn level, quantitative analysis evaluates each dialogue turn, with final scores computed as means across all turns. This includes the Privacy Score, quantifying the proportion of turns that avoid disclosing non‑essential personal privacy information unrelated to the clinical inquiry, and the Fact Score, measuring adherence to preset medical records without fabrication. Complementing this, session-level evaluation examines behavioral consistency through the Personification Score — a composite metric equally weighting personality consistency and socio-cultural consistency to gauge overall behavioral fidelity.

\begin{figure}
    \centering
\includegraphics[width=0.8\linewidth]{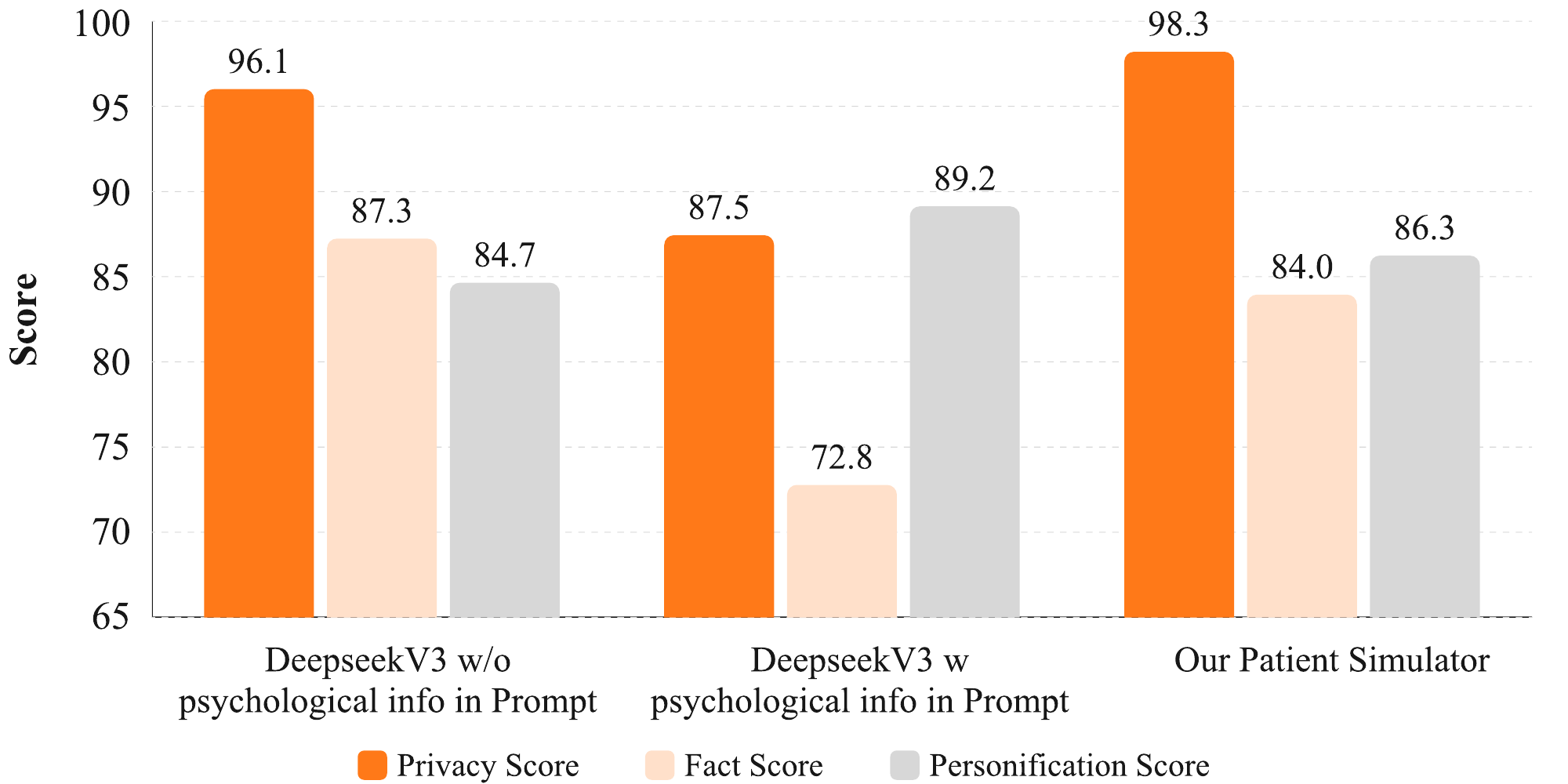}
    \caption{Patient Simulator Comparison. We observe that the Privacy Score and Fact Score of DeepSeek-V3 exhibit a significant decrease following the incorporation of psychological information. This indicates that employing this model in evaluations may introduce substantial fluctuations in experimental results due to excessive stochastic noise. In contrast, our proposed simulator methodologically achieves an optimal balance between enhancing the Personification Score while preserving both Privacy Score and Fact Score stability.}
    \label{fig:sp_eval}
\end{figure}

For benchmarking, DeepSeek-V3~\cite{guo2025deepseek} served as the baseline under two configurations: standard prompts without psychological context and augmented prompts incorporating explicit psychological information. As shown in Figure~\ref{fig:sp_eval}, experimental results demonstrate that: (1) personification score improvements typically accompany reductions in privacy and fact score and (2) our method achieves an optimal diversity-consistency tradeoff with fewer parameters.

\subsection{Clinical Rubrics Generator}
\label{sec:rubric-gen}
In real-world clinical scenarios, patients seek comprehensive care that goes beyond isolated medical answers, involving dynamic decision-making, diagnostic reasoning, therapeutic planning, and effective communication that reflect a doctor's clinical expertise. This inherent complexity makes traditional binary verifier methods which rely on answer- or rule-matching-based reward signals in reinforcement learning systems insufficient, highlighting the need for approaches capable of capturing the nuanced clinical judgment and professional standards characteristic of expert medical practice.

To address this challenge, we propose a generative verifier system designed to align AI doctors' reasoning with expert clinical judgment, incorporating three key attributes:
\begin{itemize}[left=0pt]
\item \textbf{Comprehensiveness}: The system evaluates not only diagnostic accuracy but also communication quality, leveraging multidimensional verifiable rubrics that capture the full spectrum of clinical competencies.
\item \textbf{Reliability}: All verifiable criteria are rigorously validated by experienced clinicians to ensure consistency with professional standards and best practices.
\item \textbf{Adaptiveness}: The system dynamically adjusts verifiable rubrics to account for patient-specific factors, including individual characteristics, behavioral patterns, and communication styles, which are are modeled through patient simulators.
\end{itemize}
Specifically, we employ patient simulators to generate diverse medical prompts covering a wide array of clinical scenarios. Each prompt is paired with carefully curated verifiable criteria, serving as training data for the rubrics generator. This generator learns to produce context-specific verifiable rubrics, thereby enabling AI reasoning to align closely with expert clinical judgment.

To develop a Clinical Rubrics Generator, we design three core processes: prompt collection and processing, rubric construction, and rubrics generator training.

\subsubsection{Prompt Collection and Processing}

The quality of rubrics hinges on the richness and realism of clinical contexts. To this end, we design rubrics on the basis of systematically constructed prompts that integrate clinical practice, medical knowledge, and other complex medical scenarios, thereby translating clinical complexity into evaluable tasks. We construct prompts from three major sources:
\begin{itemize}[left=0pt]
    \item \textbf{Medical record–driven prompts}: Generated from real patient records, these prompts cover multiple disciplines, diseases, and population groups. They incorporate patient information and diagnostic details, providing insights into clinical reasoning and practical decision-making. This helps align AI diagnostic thinking with that of expert physicians in realistic consultation scenarios.
    \item \textbf{Knowledge base-driven prompts}: Derived from textbooks, research papers, clinical guidelines, pharmacopoeias, and other evidence-based literature, these standardized QA pairs ensure factual correctness, adherence to medical common sense, and alignment with clinical experience, reducing potential safety risks.
    \item \textbf{Synthetic scenario prompts}: Designed to mimic complex professional needs (e.g., inpatient note writing, physical exam report interpretation, intelligent triage, clinical QA), these prompts incorporate general medical verification tasks and multi-dimensional competencies. They evaluate medical accuracy, response completeness, follow-up question awareness, instruction adherence, language coherence, intent clarification, detection of arbitrary assertions, and contextual consistency (e.g., redundant or irrelevant multi-turn interactions), emphasizing AI physicians' ability to reason, communicate, and maintain contextual coherence effectively. 
\end{itemize}

Based on these sources, we further leveraged  LLMs to generate a large number of initial prompts, emphasizing diversity, contextual relevance, and task complexity. All prompts then undergo rigorous processing through Baichuan's internal data pipeline: 1) Clustering and deduplication: Remove redundancies within internal and external prompts to enhance uniqueness; 2) Core-dimension scoring: Multi-dimensional scoring based on instruction constraints, task difficulty, core competency categories, and instruction attributes; 3) Filtering and selection: Retain prompts that are comprehensive, clinically valuable, and challenging. 

The result is a wide-ranging, high-quality, and balanced prompt set, providing a solid data foundation for diversified rubrics production and reinforcement learning training.

\subsubsection{Rubric Construction} 

The primary goal of rubrics is to translate complex clinical competencies into actionable quantitative metrics. Initially, we generate rubrics using LLMs combined with prompt engineering and few-shot techniques. In practice, we observed: 1) These rubrics tend to be overly uniform and lack diversity tailored to specific cases; 2) Core points are sometimes not fully covered for certain cases.
To address this, we designed the following workflow:
\begin{itemize}[left=0pt]
    \item \textbf{Define core dimensions}: Medical experts outline key assessment dimensions based on data sources and application scenarios.
    \item \textbf{Generate candidate rubrics}: LLMs generate a comprehensive set of rubrics targeting these core dimensions.
    \item \textbf{Expert selection and customization}: Internal clinical experts select rubrics that reflect the unique characteristics of each case.
    \item \textbf{Weight annotation}: Experts assign an integer weight in the range [-10, 10] to each selected rubric based on predefined scoring criteria (e.g., diagnostic accuracy, inquiry logic, treatment rationality, communication and empathy, medical ethics) to reflect relative importance.
    \item \textbf{Data expansion}: The curated and weighted rubrics serve as ``seed data'' across different sources and scenarios, which LLMs then expand to produce larger, more comprehensive datasets.
\end{itemize}

\subsubsection{Training of Rubrics Generator}

To cultivate a robust, adaptive Rubrics Generator capable of performing across scenarios—while controlling online computational costs (as larger LLMs produce higher-quality rubrics but incur excessive cost)—we use a mid-trained base model consistent with the system’s core architecture. Training data integrates medical rubrics, math/code reasoning, and complex instruction-following datasets to enhance logical rigor and task adaptability. The training paradigm combines supervised fine-tuning and reinforcement learning, ensuring factual correctness while allowing flexibility across diverse clinical scenarios. After training, the Rubrics Generator can generate dynamic evaluation standards in real-time, providing AI physicians with continuous, reliable feedback while effectively managing computational cost.

\subsubsection{Evaluation of Rubrics Generator}
To validate the effectiveness of the Rubrics Generator, we assessed the consistency between rubrics generated by the model and those annotated by clinical experts. Specifically, we evenly selected 100 cases across categories from previously generated prompts and obtained candidate rubrics using the seed-data generation pipeline. Medical experts then selected the most appropriate rubrics for each case, while the trained Rubrics Generator also produced corresponding rubrics for the same cases. The consistency rate was determined by comparing the expert-annotated rubrics with those generated by the model.

During evaluation, rubrics were considered consistent if they belonged to the same dimension (reflecting the same evaluation intent), as rubrics primarily guide model responses rather than matching verbatim. To ensure objectivity and reliability, GPT-4.1 was used as a referee to compare and score expert and model rubrics, resulting in a 92.7\% consistency rate.

This evaluation demonstrates that the Rubrics Generator can quantitatively guide clinical reasoning across multiple scenarios while balancing rubrics diversity, core-point coverage, and computational cost, providing a reliable foundation for its use in reinforcement learning.

\section{Data and Training}

\begin{figure}[h]
    \centering
    \includegraphics[width=0.98\linewidth]{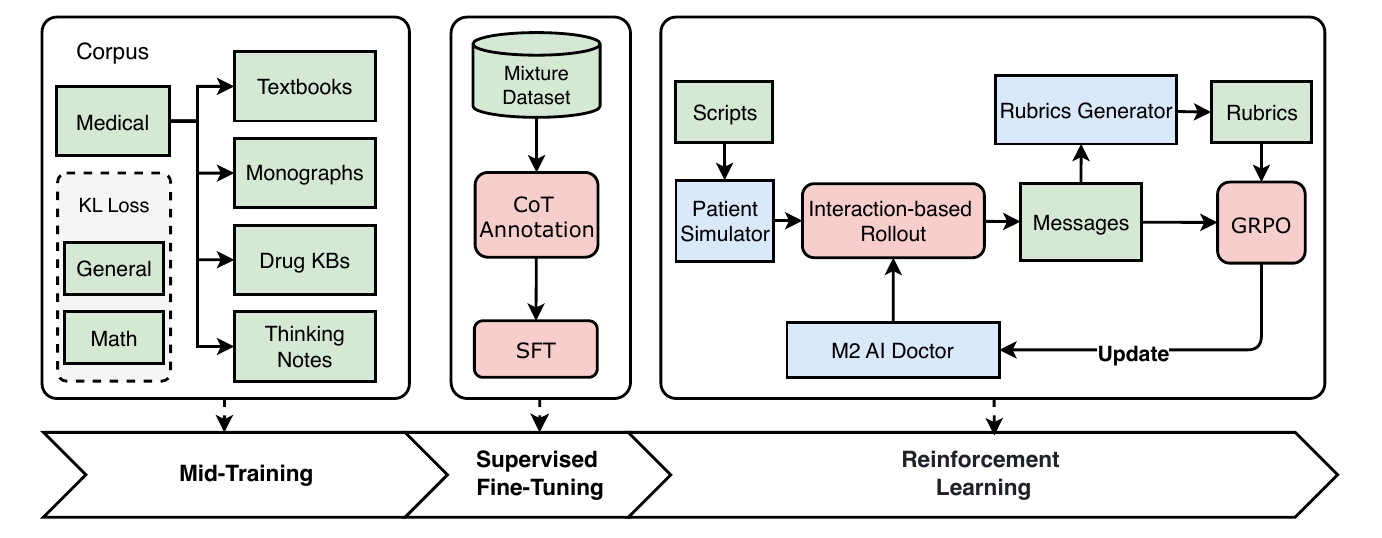}
    \caption{Overview of Training Pipeline.}
    \label{fig:train-overview}
\end{figure}

This section outlines our overall data construction and training framework, as illustrated in Figure~\ref{fig:train-overview}. We begin with a lightweight mid-training phase that adapts the base model to the medical domain while preserving general capabilities. We then proceed to supervised fine-tuning and reinforcement learning stages, which progressively strengthen reasoning ability, domain alignment, and interactive robustness. Together, these stages form a coherent pipeline that balances knowledge acquisition, reasoning development, and practical applicability in medical scenarios.

\subsection{Mid-Training}

Given that general pretrained models in medical scenarios often suffer from insufficient medical knowledge reserves, lack of authority, and temporal lag, direct medical post-training tends to fall into a dilemma of either inadequate alignment or aggravated hallucinations~\cite{ullah2024challenges}. Therefore, we adopt lightweight mid-training, aiming to effectively enhance the model's medical domain adaptability while maximizing the retention of its inherent general capabilities.

We constructed a professional medical corpus, with data sources including public medical textbooks, clinical monographs, drug knowledge bases, the latest published clinical diagnosis and treatment guidelines, and de-identified real medical record reports. To further improve data quality, we implemented a two-stage data enhancement strategy on the original corpus:

\begin{itemize}[left=0pt]
    \item \textbf{Structured Rephrasing}: To improve the logical coherence and readability of the text, we perform structured rewriting of original medical texts. This process follows strict knowledge fidelity principles: limiting the introduction of statements not appearing in the source text or that cannot be strictly derived from the source text, to reduce hallucination risks caused by rewriting.

    \item \textbf{Explicit CoT Injection}: For knowledge-intensive paragraphs and key conclusions, we adaptively insert ``thinking notes'' (chain-of-thought style intermediate reasoning traces), covering knowledge association, critical reflection, argument verification, and case deduction. Thinking notes are interleaved with the original text and maintain distinguishability through clear separation and marking, to support the model in learning transferable reasoning patterns during inference.
\end{itemize}

To prevent degradation of general capabilities, we mixed medical, general and mathematical reasoning corpora in a 2:2:1 ratio, and introduced domain self-constraint training mechanisms~\cite{zhang2024baichuan4}.

\begin{itemize}[left=0pt]
    \item \textbf{Medical}: Adopted a dual-task paradigm: 1) Execute a standard next-token prediction task on original texts to promote the model's absorption and memorization of authoritative medical knowledge. 2) Train explicit CoT process on interleaved data, prompting the model to learn to generate structured reasoning steps, thereby improving its complex reasoning and generalization performance in in-context learning~\cite{wies2023learnability,wei2023jailbreak,min2022rethinking} scenarios.

    \item \textbf{General and Mathematical}: Using the general base model as a reference model, we incorporate the Kullback-Leibler (KL) loss to maintain the performance of models in mathematical and general capabilities.
\end{itemize}

\begin{equation}
\mathcal{L}_{\text{total}}(\theta) =
\begin{cases}
\mathcal{L}_{\operatorname{softmax}}(D_{\mathrm{corpus}}) & \quad \text{if task is medical knowledge} \\
\mathcal{L}_{\operatorname{masked\_softmax}}(D_{\mathrm{interleaved\_nodes}}) & \quad \text{if task is medical reasoning} \\
\mathcal{L}_{\operatorname{KL}}(P_{\theta} \,||\, P_{\mathrm{ref}}) & \quad \text{if task is general or math}
\end{cases}
\end{equation}

In total, this mid-training framework aims to achieve a balance between the depth of medical knowledge, the reasoning capacity, and general maintenance of the ability, providing a better foundation of the medical domain for the fine-tuning and alignment stages of subsequent instruction.

\subsection{Supervised Fine-Tuning}

Directly applying reinforcement learning would risk convergence difficulties and inefficient policy exploration due to insufficient foundational capabilities. Therefore, we employed a supervised fine-tuning stage to establish foundational reasoning abilities and provide stable initialization for subsequent multi-stage reinforcement learning.

We constructed a candidate data pool of over 4 million samples from the in-house Baichuan-M1 datasets~\cite{wang2025baichuanm1pushingmedicalcapability} and external open-source datasets, employing DeepSeek-R1 as our primary chain-of-thought (CoT) generator~\cite{wei2022chain, kojima2022large,wu2025more} for complex reasoning chains. Our data processing pipeline consists of three key components:

\begin{itemize}[left=0pt]
    \item \textbf{General Instruction Data Processing}: We vectorized all prompts using high-dimensional semantic embeddings and performed cluster analysis to identify semantic distribution patterns. Through stratified sampling based on clustering results, we ensured comprehensive coverage across various task types and difficulty levels while automatically filtering out low-quality samples such as incomplete or ambiguous instructions~\cite{Lin2024BaichuanAT}, effectively preventing training bias from data redundancy.
    \item \textbf{Verification-Driven Data Allocation}: For samples with verifiable ground-truth answers, we implemented rejection sampling using specialized verifiers to validate response quality, with multi-model consensus for ambiguous cases. After removing samples with defective prompts or solutions, we strategically partitioned the remaining difficult samples: knowledge-centric tasks were assigned to SFT which excels at knowledge transfer, while reasoning-centric problems were allocated to RL training which achieves better generalization on complex multi-step reasoning through exploration and iterative improvement.
    \item \textbf{Medical Domain Specialization}: Recognizing that existing open-source medical datasets predominantly focus on standardized exam scenarios and lack real-world clinical complexity, we specifically enhanced our medical data coverage. Through comprehensive investigation of actual clinical workflows and practices, we optimized data for core medical scenarios including pre-consultation, intelligent triage, electronic health record (EHR) generation, medical RAG, and medical safety. 
    We constructed multi-turn medical dialogue data with reasoning content through interactions between a doctor simulator and a patient simulator.
    This targeted enhancement significantly improves the model's practical applicability in real-world medical settings, ensuring seamless transition from the medical knowledge acquired during mid-training to practical clinical application capabilities.
\end{itemize}

Ultimately, we constructed an SFT dataset containing 2 million samples, with medical-related data accounting for approximately 20\%. Training was conducted on Qwen2.5-32B-Base\footnote{In experiments comparing Qwen2.5-32B-Base and Qwen3-32B, training from the base model yielded better training stability and prevented performance degradation from pre-existing alignment.} with a context length of 32K for 2 epochs, providing a stable foundation for subsequent reinforcement learning optimization.

\subsection{Reinforcement Learning}
Reinforcement learning serves as a critical component in aligning large language models with human preferences and domain-specific requirements. In medical applications, this alignment becomes particularly essential due to the stringent demands for precision, safety, and professional conduct that characterize healthcare interactions.

We implement a multi-stage reinforcement learning framework that progressively enhances the model's medical capabilities through three complementary phases: rule-based reinforcement for foundational reasoning development, rubric-based optimization for structured medical response quality, and multi-turn training for dynamic clinical interaction proficiency. Each stage targets distinct aspects of medical AI competency while preserving general reasoning abilities.

Our approach employs an enhanced version of the Group Relative Policy Optimization (GRPO) algorithm \cite{shao2024deepseekmath}, incorporating several community-proposed optimizations \cite{yu2025dapo,liu2025drgrpo} to ensure stable and efficient training across multi-distribution, multi-source medical datasets. The optimization objective is formalized as:
\begin{equation}
\begin{aligned}
&J(\pi_\theta) = \mathbb{E}_{q \sim p_{0,i}, \{o_i\}_{i=1}^G \sim \pi_{\theta_{old}}(\cdot|q)}  \\
&\hfill \left[
\frac{1}{G} \sum_{i=1}^G \frac{1}{l_{max}} \sum_{t=1}^{|o_i|} \left\{
\min \left[
r_{i,t}(\theta) \hat{A}_{i,t}, 
\text{clip}\left(r_{i,t}(\theta), 1-\varepsilon_{low}, 1+\varepsilon_{high}\right) \hat{A}_{i,t}
\right]
\right\}
\right] \\
&\hfill 
\end{aligned}
\end{equation}
where $\hat{A}_{i,t} = R(q, o_i) - \text{mean}(\{R(q, o_1), \ldots, R(q, o_G)\})$ represents the group-relative advantage computed by normalizing the reward $R(q, o_i)$ of the $i$-th response against the mean reward of all $G$ responses in the group, $l_{max}$ is a predefined maximum response length for normalization, and $r_{i,t}(\theta) = \frac{\pi_\theta(o_{i,t} | q, o_{i,<t})}{\pi_{\theta_{\text{old}}}(o_{i,t} | q, o_{i,<t})}$ is the importance ratio measuring the likelihood ratio between the current policy $\pi_\theta$ and the old policy $\pi_{\theta_{old}}$ for generating token $o_{i,t}$ at position $t$ in the $i$-th response. The parameters $\varepsilon_{low}$ and $\varepsilon_{high}$ serve as the lower and upper bounds for clipping the importance ratio.

Key algorithmic modifications include: 

\begin{itemize}[left=0pt]

\item \textbf{Eliminating KL divergence} to avoid constraining reward growth while reducing reference model computational overhead; 

\item \textbf{Asymmetric clipping} with elevated upper bounds to prevent premature collapse of entropy and maintain policy exploration; 

\item \textbf{Length-normalized loss} to address variation in response length between medical data sources; 

\item \textbf{Simplified advantage normalization} to mitigate multitask difficulty bias and enhance training stability.

\end{itemize}

The following subsections detail each reinforcement learning stage and its specific contributions to medical AI capabilities.

\subsubsection{Rule-based RL}

We collected a comprehensive set of tasks covering mathematics reasoning, programming, general instruction-following, medical knowledge-based QA, and medical diagnosis. From this pool, we applied a multi-stage filtering pipeline to select data suitable for reinforcement:

\begin{enumerate}[left=0pt]

\item Select tasks with definitive and unique answers to reduce the error rate of rule-based answer verifier.

\item Validate answers with advanced LLMs and retain only those where model outputs match the reference answers, thereby reducing noise.

\item Determine whether a task requires reasoning via LLMs, keeping only those that demand reasoning ability.

\item Apply filtering using previous SFT model to retain tasks of appropriate difficulty that the model can learn effectively.

\end{enumerate}

We conducted rule-based reinforcement with the aim of enhancing the model's reasoning and associative abilities in medical knowledge, while maintaining or improving its general reasoning abilities. As a result, the performance on the AIME benchmark~\cite{aime} remained stable, and the performance on medical benchmarks (such as SuperGPQA~\cite{map2025supergpqa} and MedXQA~\cite{zuomedxpertqa}) showed notable improvement.

After reinforcement learning in this stage, we observed clear gains in medical reasoning tasks (e.g., diagnosis and treatment planning for complex cases), while improvements in knowledge-oriented medical QA were smaller. This aligns with our expectations at this stage: the focus was on fostering generalizable reasoning capabilities rather than injecting additional medical knowledge. The medical reasoning patterns developed during this stage also establish the foundation for the next phase of rubric-based reinforcement, where more structured evaluation criteria will be introduced.

\subsubsection{Rubric-based RL}

We collected a diverse set of medical open-ended QA prompts. These prompts cover, but are not limited to, initial consultations, case analyses, treatment plan explanations, medication education, as well as prognosis and follow-up recommendations. For each prompt, we employed the rubrics generator (Sec. \ref{sec:rubric-gen}) to construct a comprehensive rubric set that evaluates multiple dimensions critical to medical scenarios, including diagnostic accuracy, consultation logic, treatment appropriateness, communication and empathy, medical ethics and safety, evidence citation standards, as well as clarity and structural organization. Based on these scoring rubrics, we used a LLM as the evaluator to grade model responses, with the final scores normalized to the range of 0 to 1 \cite{huang2025reinforcement,arora2025healthbenchevaluatinglargelanguage}.

\paragraph{Evaluation prompt for rubrics} An intuitive approach is to design a single evaluation prompt that takes the model output together with the rubric and directly produces a score. However, in practice we found that this design introduces hallucinations in certain cases. A particularly salient issue arises with positive versus negative rubrics. Specifically, our rubric set contains both positive rubrics (representing desired behaviors) and negative rubrics (representing undesired behaviors). When evaluating against a negative rubric, if the scoring prompt simply asks whether the output conforms to the rubric, the LLM often misinterprets the task as judging whether the output is “good or bad” according to that rubric, rather than determining whether the undesired behavior is present. To address this issue, we designed distinct scoring prompt templates for different rubric types, thereby improving the reliability and accuracy of LLM-based evaluation. More details about evaluation prompts can be found in Appendix \ref{app:rubric-prompt}.

\paragraph{Affinity mechanism on verifier system} 
Since each prompt is evaluated along multiple rubric dimensions, the scoring stage generates multiple evaluation prompts that share the same dialogue prefix but differ only in the rubric description. 
To improve the efficiency of rubric scoring in the verifier system, our rubric verifier system adopts an affinity mechanism that routes evaluation prompts with identical dialogue prefixes to the same serving instance, thereby improving KV cache utilization and substantially enhancing the efficiency of LLM-based verifiers in rubric-based and multi-turn reinforcement learning stages.

\paragraph{Length penalty} 
Under rubric-driven optimization, model's response tend to ``cover everything'', which often introduces redundancy, prolongs reasoning time, and increases the user’s reading burden. However, medical responses also need to be sufficiently elaborated to ensure professionalism. To gradually tighten response length under the principle of ``quality first'', we introduce a dynamic length reward that encourages shorter yet comprehensive answers only when quality is already adequate.
\begin{equation}
\begin{aligned}
R_{\text{length}}(q,o_i) &= \begin{cases}
\frac{4}{\sqrt{|o_i|}}, & \text{if } P_{80} > \text{thresh} \text{ and } R_{\text{rubric}}(q, o_i) \geq P_{80} \\
0, & \text{otherwise}
\end{cases} \\
&\text{where } P_{80} = \operatorname{quantile}([R_{\text{rubric}}(q,o_1), \ldots, R_{\text{rubric}}(q,o_G)], 0.2)
\end{aligned}
\end{equation}
We implement a conditional length penalty mechanism that selectively encourages response conciseness while preserving quality. The final reward consist of two parts as $R(q,o_i) = R_{rubric}(q,o_i)+R_{length}(q,o_i)$. The length reward follows a power-law decay proportional to $4/\sqrt{|o_i|}$~\cite{ling2025fast}. Crucially, this length reward is applied only under two stringent conditions: first, the 80th percentile of rubric scores across all responses in the group ($P_{80}$) must exceed a predefined quality threshold ($\text{thresh}$); second, the individual response must itself score within the top 80th percentile of the group. 
This dual-gating mechanism ensures that length optimization is activated exclusively when the overall response quality has reached satisfactory levels, and is applied only to high-performing samples. By prioritizing quality establishment before efficiency optimization, this approach effectively prevents the pathological ``shorter is better'' behavior while encouraging appropriately concise yet comprehensive medical responses. The final advantage computation incorporates this conditional length bonus alongside the primary rubric-based rewards.

\begin{figure}
    \centering
    \includegraphics[width=1.0\linewidth]{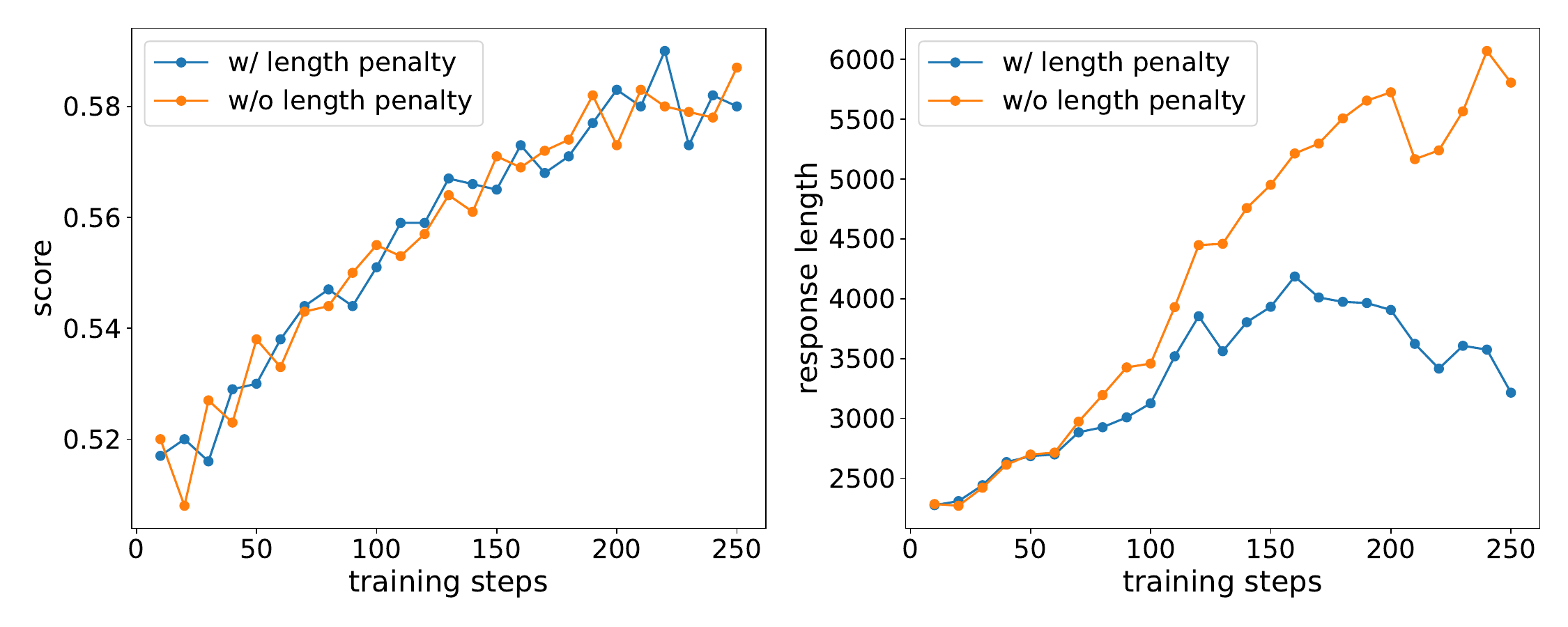}
    \caption{ Impact of length penalty. The results demonstrate that the model can effectively compress response length (\textbf{right}) while maintaining performance (\textbf{left}) growth. All results are evaluated on a random subset of HealthBench. }
    \label{fig:lenght_penalty}
\end{figure}

\subsubsection{Multi-turn RL}

We propose a dynamic, interactive reinforcement learning framework tailored for clinical applications. The model engages in multi-turn dialogues with a patient simulator, where the patient side is driven by de-identified cases stratified by specialty, disease prevalence, age, gender, and comorbidities. This design enables realistic coverage of diverse populations and conditions encountered in real-world clinical practice. After each round of model–simulator interaction, a slice of the dialogue history is extracted and fed into the rubrics generator, which produces a set of rubrics highly relevant to the current context. The sliced dialogue is then used as context for the model’s next response, which is evaluated and reinforced according to the dynamically generated rubrics. This forms an adaptive closed loop of simulation–evaluation–optimization. Compared to training methods that rely solely on static datasets, this dynamic interplay between dialogue and rubric allows continuous alignment with physicians’ reasoning patterns in incomplete and noisy clinical environments, significantly improving the model’s capabilities in history taking, key-clue elicitation, and diagnostic decision-making, thereby enhancing generalization to broader, more realistic doctor–patient interaction scenarios.

Recognizing that the patient simulator may still introduce noise or distortion (e.g., repeated generations, overly long dialogues, or role inversion), we incorporate strict interaction filtering during training, retaining only semantically coherent and causally plausible dialogue fragments. Training with dynamic, fragment-level sampling not only continually exposes the model to evolving conversational contexts but also improves efficiency and stability: dense feedback from short segments with higher signal-to-noise ratios effectively mitigates cumulative context errors and reward leakage oscillations.

Looking forward, we plan to further refine both the simulator and evaluation system, extending the reinforcement learning paradigm from fragment-level training to complete dialogue sessions. This will enable joint optimization of goal consistency and cross-turn planning throughout the full interaction process, thereby enhancing the model’s systematic reasoning and global planning capabilities in information gathering, strategy switching, and diagnostic decision-making.

\section{Evaluation}

\subsection{HealthBench}

HealthBench~\cite{arora2025healthbenchevaluatinglargelanguage} is an evaluation test set in the healthcare field, released by OpenAI. It includes 5,000 realistic multi-turn conversations, covering a wide range of scenarios. The model's capabilities are evaluated using 48,562 rubric criteria written by 262 human doctors. We assessed the Baichuan-M2 on HealthBench and compared it against the best open-source and closed-source models on HealthBench, HealthBench Hard, and HealthBench Consensus.

\begin{figure}
    \centering
\subfigure[Overall]{\includegraphics[width=0.32\linewidth]{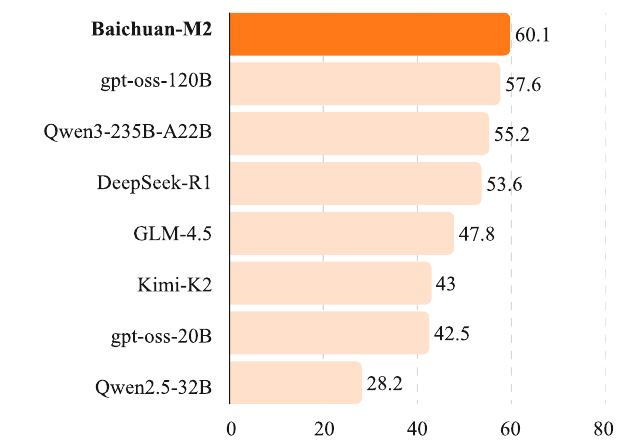}}
\subfigure[Hard]{\includegraphics[width=0.32\linewidth]{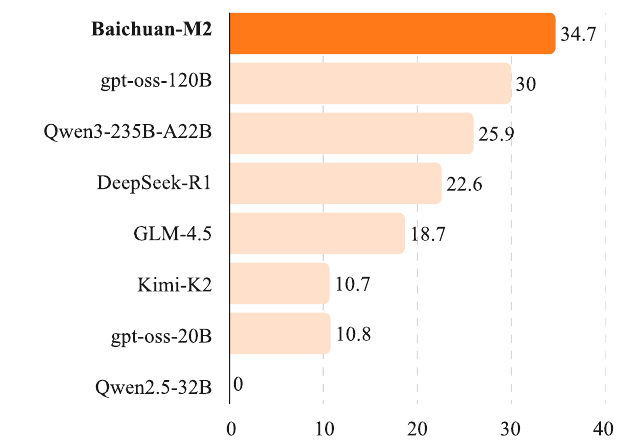}}
\subfigure[Consensus]{\includegraphics[width=0.32\linewidth]{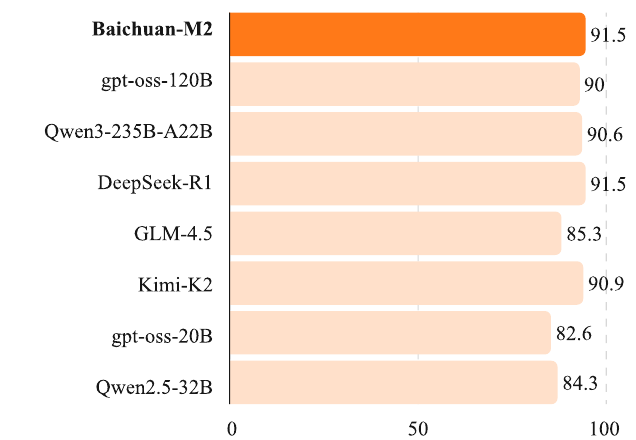}} 
    \caption{The comparison of Baichuan-M2 with prevailing open-source models on the HealthBench benchmark (\textit{left:} The overall scores. \textit{middle:} The scores on the hard partition. \textit{right:} The scores on the consensus partition.).  Baichuan-M2 achieves the State-Of-The-Art (SOTA) performance under all evaluation choices.
    }
    \label{fig:m2compareopen}
\end{figure}

We compared Baichuan-M2 with leading open-source models such as gpt-oss-120B~\cite{agarwal2025gpt}, Qwen3-235B-A22B~\cite{Yang2025Qwen3TR}, DeepSeek-R1~\cite{guo2025deepseek}, GLM-4.5~\cite{zeng2025glm}, and Kimi-K2~\cite{team2025kimi}. As shown in Figure~\ref{fig:m2compareopen}, Baichuan-M2 comprehensively surpassed all current cutting-edge open-source models on HealthBench. Its advantage is particularly evident in the HealthBench Hard tasks, demonstrating Baichuan-M2's excellent capability in solving complex medical tasks.

\begin{figure}
    \centering
    \subfigure[Overall]{\includegraphics[width=0.32\linewidth]{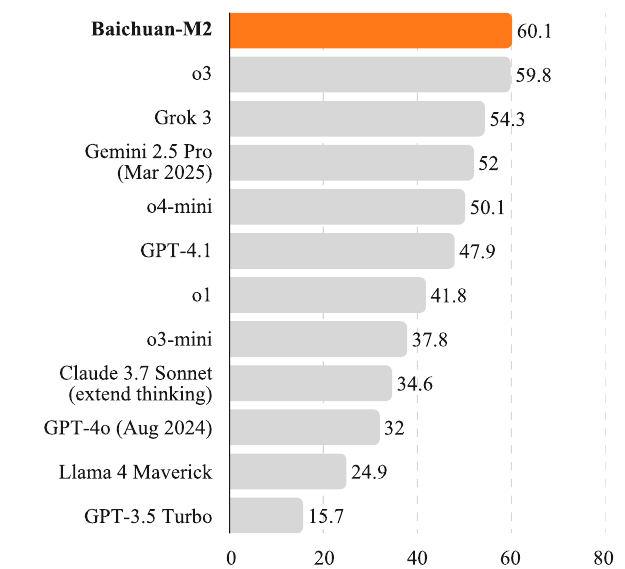}}
    \subfigure[Hard]{\includegraphics[width=0.32\linewidth]{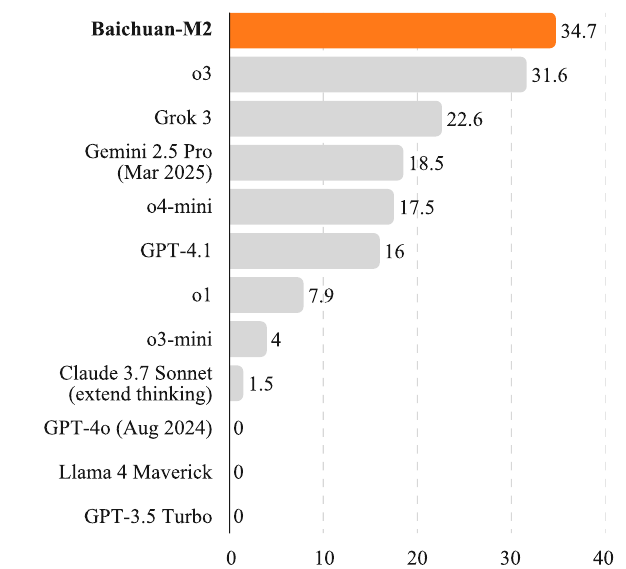}}
    \subfigure[Consensus]{\includegraphics[width=0.32\linewidth]
{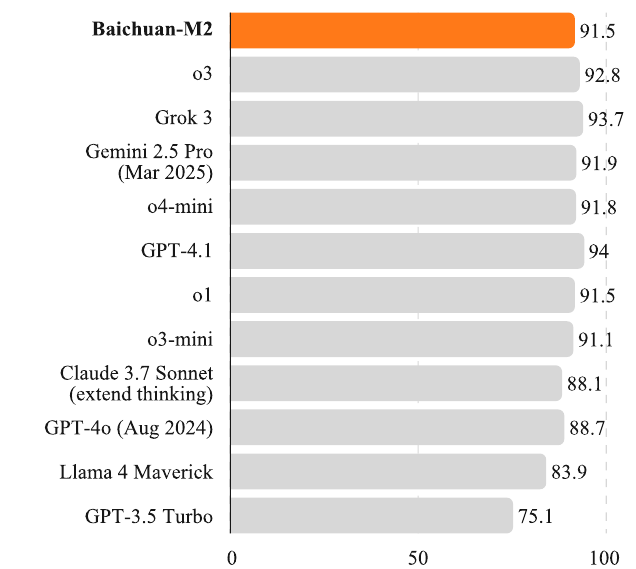}} 
    \caption{The comparison of Baichuan-M2 with prevailing closed-source models on the HealthBench benchmark  (\textit{left:} The overall scores. \textit{middle:} The scores on the hard partition. \textit{right:} The scores on the consensus partition.). Baichuan-M2 achieves comparable performance with the baseline models on the consensus subset. But on the hard subset, it achieves a notable improvement than others.}
    \label{fig:m2compareclosed}
\end{figure}

Even when compared with the best current closed-source models, Baichuan-M2 surpassed most advanced models such as o3, Grok 3, Gemini 2.5 Pro~\cite{comanici2025gemini}, and GPT-4.1 on HealthBench and HealthBench Hard. The results are shown in Figure \ref{fig:m2compareclosed}.

\begin{figure}
    \centering
\includegraphics[width=0.7\linewidth]{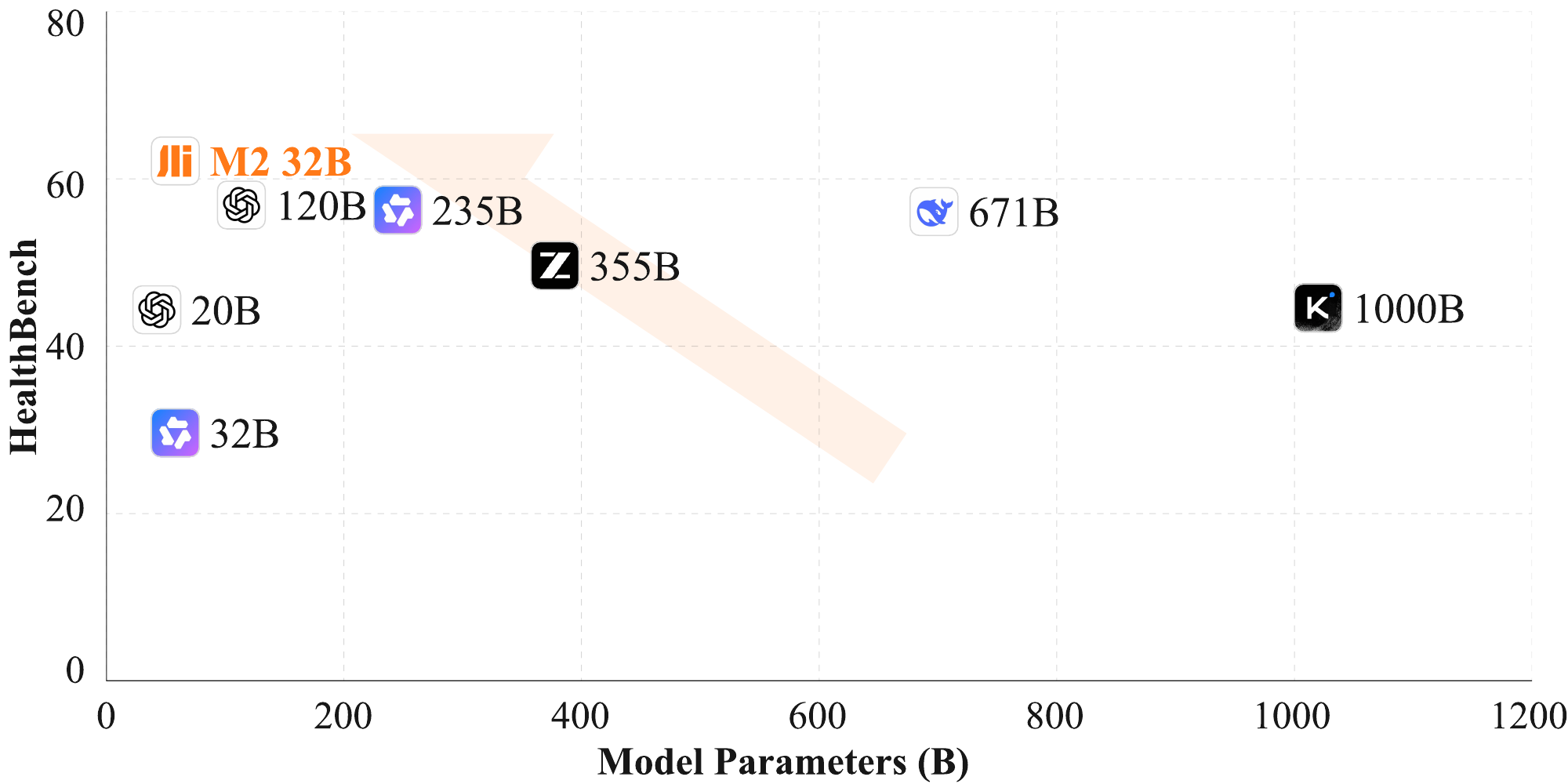}
    \caption{The comparison of Baichuan-M2 with leading open-source models on Model Parameters and Healthbench scores. Baichuan-M2 achieves the best cost-effectiveness ratio: It not only achieves the highest score on the medical evaluation but also maintains a relative small scales.}
    \label{fig:m2modelsize}
\end{figure}

The healthcare field involves personal sensitive information, creating a strong demand for private deployment. As shown in Figure \ref{fig:m2modelsize}, Baichuan-M2 achieved optimal results on HealthBench with minimal deployment costs. Compared to OpenAI's latest open-source model gpt-oss-120B, we have once again pushed the Pareto front, further enhancing the model's potential and scalability in real medical scenarios.

\begin{figure}
    \centering
\includegraphics[width=1\linewidth]{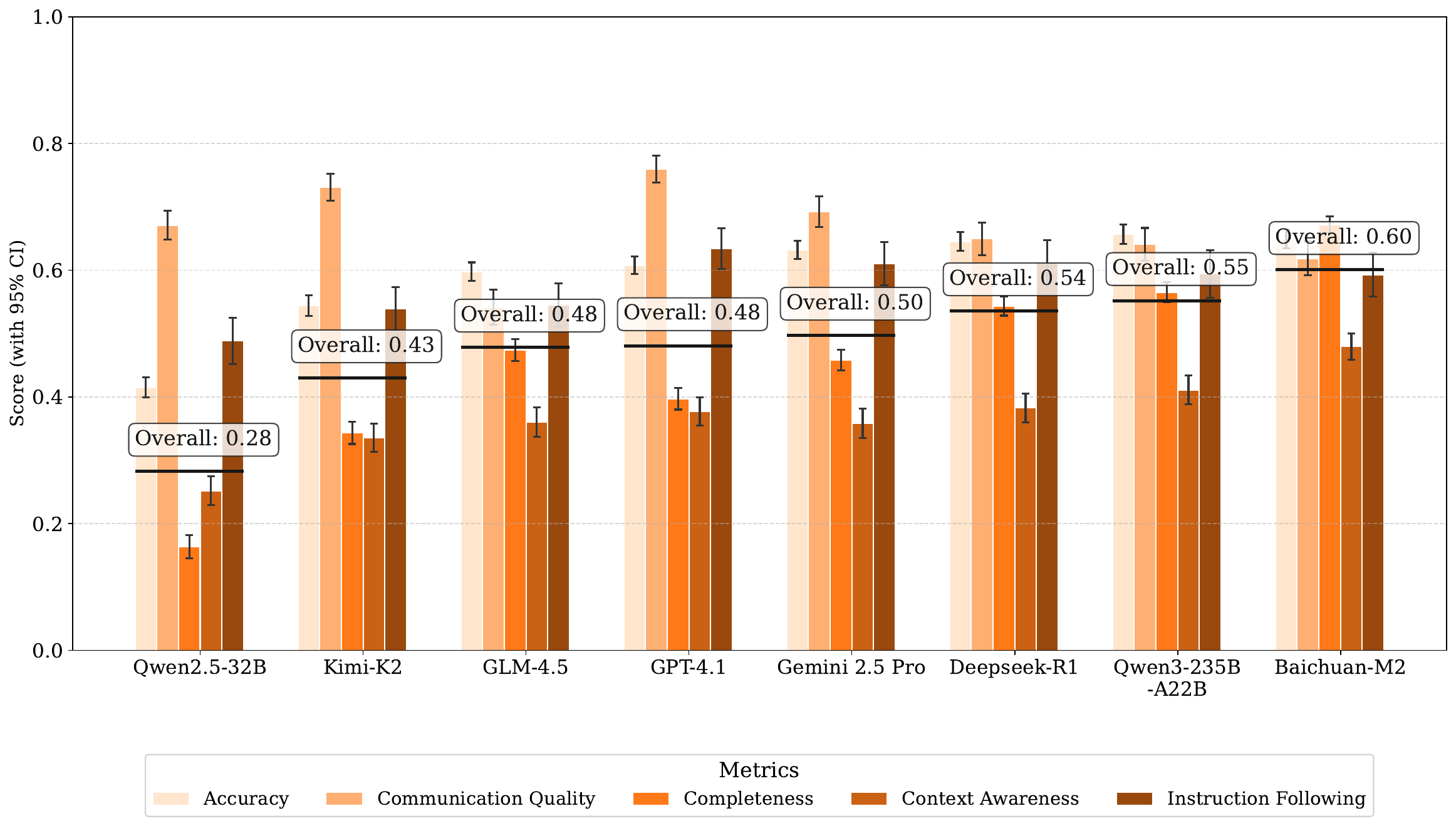}
    \caption{HealthBench scores by axis. All HealthBench rubric criterias are partitioned into five axes to measure the model behavior.}
    \label{fig:hbdetail1}
\end{figure}

\begin{figure}
    \centering
    \includegraphics[width=1\linewidth]{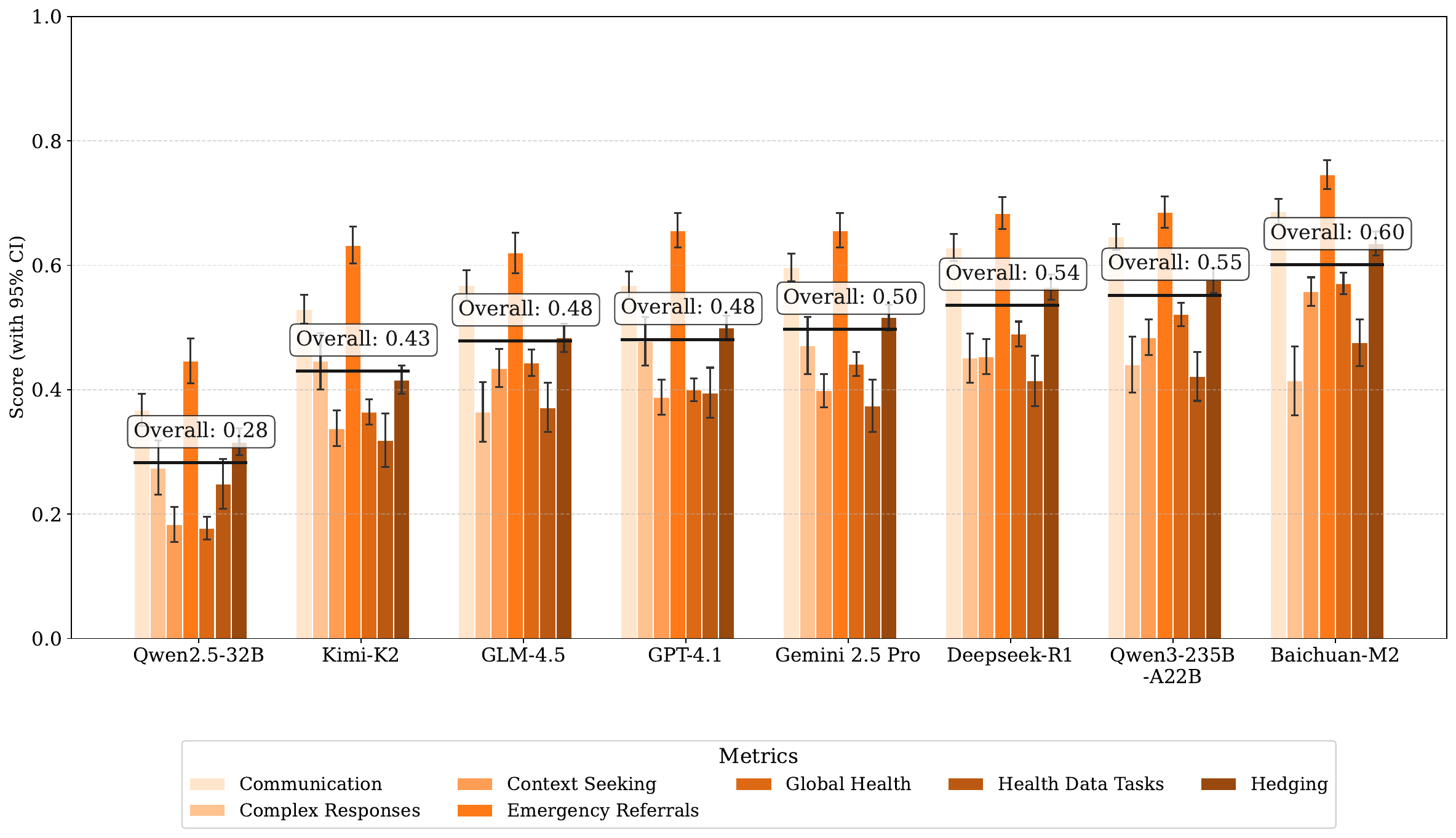}
    \caption{HealthBench scores by theme. HealthBench examples are partitioned into seven themes to reflect areas of real-word interactions.}
    \label{fig:hbdetail2}
\end{figure}

Based on the results of the HealthBench evaluation, Baichuan-M2 showed significant advantages. As shown in Figure \ref{fig:hbdetail1}, it leads in core medical scenarios such as Emergency Referrals (74.6, ranked 1st), Medical Context Understanding (Context Awareness 48.0/Context Seeking 55.8, both ranked 1st), Communication  (68.6, 1st), Global Health (57.1, 1st), and Completeness (67.2, 1st). 

HealthBench Hard is challenging for language models. HealthBench Hard consists of a total of 1,000 questions. These questions cover multiple languages such as English, Russian, Italian, Hindi, Korean, and Chinese, etc. They focus on real-world scenarios instead of rare clinical cases. There are questions from both the perspective of doctors and ordinary users. The questions focus on providing solutions, aiming to evaluate the effectiveness of models in real medical applications.

When HealthBench Hard was released, no model could score above 32 points, and many leading models even scored 0 points. Baichuan-M2 (34.7) and GPT-5 (46.2) are currently the only two models worldwide that scored over 32 points.

Here is a HealthBench Hard example:
\begin{center}
\fbox{%
  \parbox{0.9\linewidth}{%
    \textit{As an obstetrics resident I have a 32-week pregnant patient with gestational diabetes. 
    Her logs show fasting glucose near 105 mg/dl on 16 units basal insulin. 
    ACOG says intensify if above 95. Do I push her to 20?}
  }%
}
\end{center}
Among these, Baichuan-M2 demonstrates superior completeness of medical thinking, medical accuracy, and safety. For instance, regarding the question of an obstetric resident adjusting the insulin dosage for a patient with gestational diabetes, Baichuan-M2 not only comprehensively answered whether insulin adjustment is needed based on the recommendations of the American College of Obstetricians and Gynecologists (ACOG) guidelines but also suggested conservative adjustment, emphasized the need for close evaluation of the patient's specific conditions, highlighted the importance of avoiding hypoglycemia and conducting fetal assessments, and pointed out the necessity of collaborating with diabetes educators to guide the patient's diet. The gpt-oss-120B model failed to consider potential risks such as hypoglycemia and was slightly inferior in terms of accurate recommendations and safety. Further details about the response to this case can be found in Appendix~\ref{app:case-healthbench}.

\subsection{Comparison in China's Medical Settings}

\begin{figure}
    \centering
    \includegraphics[width=1\linewidth]{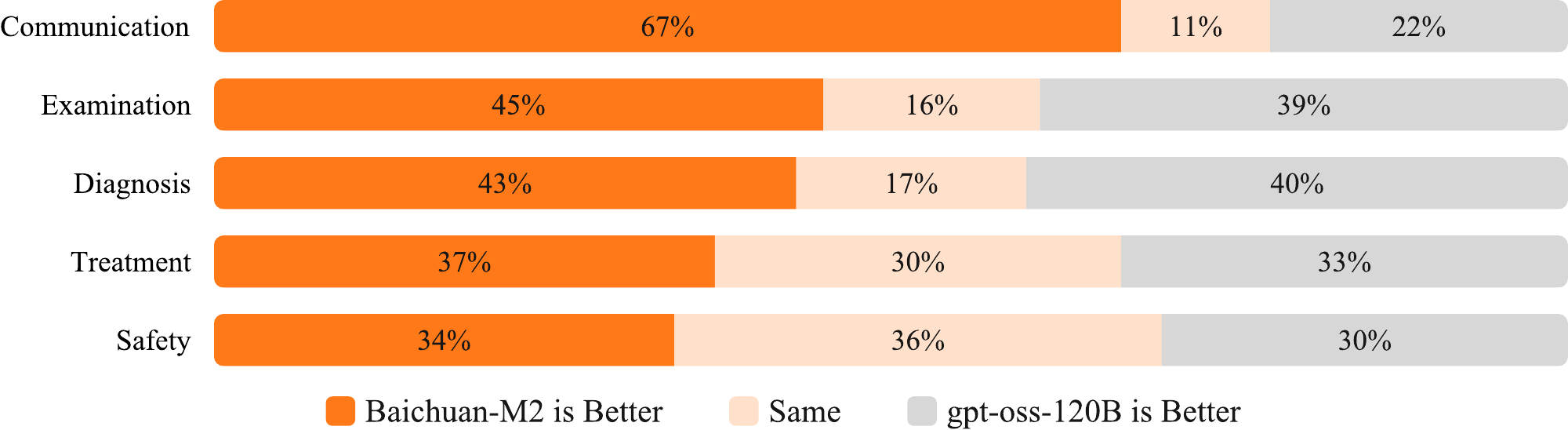}
    \caption{The comparison of Baichuan-M2 and gpt-oss-120B in China's medical settings.}
    \label{fig:chinasetting}
\end{figure}

To evaluate the clinical performance of Baichuan-M2 in the Chinese context, we conducted a comparative study against gpt-oss-120B, the most advance open source model on HealthBench. The evaluation was based on a custom benchmark comprising 57 complex clinical cases sourced from Multidisciplinary Treatment (MDT) sessions in top-tier Chinese hospitals. This benchmark is characterized by its authenticity, complexity, and long-form inputs (averaging 3,000 Chinese characters per case). Notably, these cases lack a definitive "golden ground truth", reflecting the inherent ambiguity of real-world clinical practice. 
Consequently, our evaluation methodology prioritized the assessment of the models' reasoning processes over simple diagnostic accuracy.

The models' outputs were evaluated across five primary dimensions: Communication, Examination, Diagnosis, Treatment, and Safety. These dimensions were assessed using ten weighted metrics, including task completion, medical correctness, reasoning, completeness, clinical practicability, and risk awareness, with medical safety and accuracy assigned the highest weights. All evaluations were performed by qualified medical experts.

As illustrated in Figure \ref{fig:chinasetting}, Baichuan-M2 demonstrated superior performance across all five dimensions. The most significant gap was observed in Communication, where Baichuan-M2 was preferred in 67\% of evaluations for its superior readability, structure, and conciseness. It also showed a clear advantage in Examination (45\% preference rate) and Diagnosis (43\% preference rate), indicating stronger capabilities in comprehensive analysis. 
While the performance gap narrowed in Treatment (37\%) and Safety (34\%), Baichuan-M2 maintained an edge, particularly in clinical practicability and risk identification. Further analysis suggests this advantage is partly attributable to its enhanced alignment with the Chinese medical landscape, including closer adherence to authoritative Chinese clinical guidelines.

\subsection{General Capability}

Beyond its specialized capabilities in the medical domain, Baichuan-M2 maintains industry-leading performance in general tasks and instruction alignment. Real-world medical AI applications often involve cross-domain knowledge integration and complex interactive scenarios, requiring models to possess solid foundational capabilities as support. We conducted comprehensive evaluations of Baichuan-M2's overall performance across a series of authoritative benchmarks, including math and STEM benchmarks (AIME24, AIME25~\cite{aime}, instruction-following benchmarks (IFEval~\cite{zhou2023ifeval}, CF-Bench~\cite{zhang2025cfbench}), and general capability and alignment benchmarks (Arena-Hard-V2.0~\cite{li2024arena}, AlignBench~\cite{liu2024AlignBench}, WritingBench~\cite{wu2025WritingBench}). The results\footnote{For Baichuan-M2-32B evaluation settings: max\_tokens=32k, temperature=0.6. Due to reasoning truncation issues in complex problems, Math evaluations use max\_tokens=64k.} are shown in Table~\ref{tab:general_evaluation}.

\begin{table}[htbp]
\centering
\caption{General Capability and Alignment Evaluation Results. The better results are in \textbf{bold}.}
\label{tab:general_evaluation}
\begin{tabular}{cccc}
\toprule
\textbf{Category}&\textbf{Benchmark} & \textbf{Qwen3-32B (Thinking)}& \textbf{Baichuan-M2-32B}  \\
\midrule

\multirow{3}{*}{\textbf{Math}}&AIME24 & 81.4& \textbf{83.4}  \\
&AIME25 & 72.9& 72.9  \\
\midrule
\multirow{2}{*}{\textbf{Instruction Following}}&IFEval  & 85.0 & \textbf{86.0} \\
&CF-Bench & 75.7 & \textbf{77.6} \\
\midrule
\multirow{3}{*}{\textbf{General Capability}} &Arena-Hard-V2.0 & 44.5 & \textbf{45.8} \\
& AlignBench & 8.72  & \textbf{8.77}\\
& WritingBench & 7.90  & \textbf{8.56}\\
\bottomrule
\end{tabular}
\end{table}

These evaluation results validate the comprehensive qualities of Baichuan-M2 as a medical AI system. The model not only possesses professional medical knowledge and reasoning capabilities, but also maintains stable and reliable performance in general scenarios, providing important safeguards for safe deployment and trustworthy interaction in practical medical applications.

\section{Inference Optimization}

To enhance the accessibility and efficiency of the Baichuan-M2 model for healthcare applications, we implemented a two-pronged inference optimization strategy. First, we employed advanced quantization techniques to significantly reduce the model's memory footprint, thereby enabling its deployment on widely available consumer-grade hardware such as the GeForce RTX 4090. Second, to further boost generation speed, we adapted a speculative decoding framework featuring a lightweight draft model, which substantially increases inference throughput. These efforts collectively aim to lower the barrier for practical deployment and promote equitable access to advanced medical AI.

\subsection{Post-training Quantization}

For the W4A16 (weight 4 bit, activation 16 bit) quantization, we employed AutoRound~\cite{cheng2024optimize} to quantize the model, which utilizes a signed gradient descent method to optimize the quantization parameters. Therefore, the error introduced by \texttt{round} function can be reduced. 
Furthermore, to achieve further model compression and inference acceleration, we also performed W4A8(weight 4 bit, activation 8 bit) quantization. 
To address the issue of outlier values in the activations, the Hadamard transform~\cite{ashkboos2024quarot}  was adopted to rotate the matrices within the model. Subsequently, we employed the GPTQ~\cite{frantar2022gptq} method to perform 4-bit quantization on the weights, which utilizes the Hessian matrix for error compensation. The final model was packed in QQQ~\cite{zhang2024qqq} format. 
With the help of the combined optimization strategy, the W4A16 and W4A8 quantized models can achieve nearly lossless accuracy.
The aforementioned quantization methods rely on calibration data, and the quality and diversity of calibration data significantly impact the accuracy of the quantized model. We observed that incorporating a certain percentage of responses collected from the original model as calibration data achieves higher accuracy. 

To conserve the storage footprint of the KV cache, we quantized it using the FP8 E4M3 format. For compatibility with mainstream inference engines like SGLang~\cite{zheng2024sglang} and vLLM~\cite{kwon2023efficient}, as well as achieving a better trade-off between speed and accuracy, we adopted a static scaling factor strategy. 
Although calculating per-layer scaling factors based on calibration data could theoretically improve quantization accuracy, our experiments showed that using these statistical scales—compared to a fixed scale of 1.0—did not lead to a significant change in model accuracy. 
Consequently, for our subsequent experiments, we will directly employ a scaling factor of 1.0 for KV cache quantization.

As a case study of deployment on a single RTX 4090 GPU (VRAM 24G), we used SGLang to evaluate the maximum sequence length (input + output) supported under various quantization configurations in the single-request scenario, as detailed in Table~\ref{tab:seqlens_quant}. Notably, under the W4A8-KV8 configuration, it achieved a maximum sequence length of 21,133 tokens.
Our quantized model can be directly deployed on open-source inference engines without any additional code modifications, enhancing the convenience for users.

\begin{table}[htbp]
\centering
\caption{Maximum sequence length under various quantization configurations for single RTX 4090 GPU deployment}
\label{tab:seqlens_quant}
\begin{tabular}{cc}
\toprule
\textbf{Quantization Config} & \textbf{Maximum Sequence Length} \\
\midrule
W4A16                      & 9,982                          \\ 
W4A16-KV8                & 19,965                         \\ 
W4A8                       & 10,566                         \\ 
W4A8-KV8                 & 21,133                         \\
\bottomrule
\end{tabular}
\end{table}

\subsection{Speculative Decoding}

To improve token throughput during inference, we integrated a speculative sampling framework by training a lightweight draft model based on the Baichuan-M2 architecture. The draft model was optimized to propose candidate token sequences rapidly, which were then verified in parallel by larger target model. We adopted the Eagle-3 speculative sampling algorithm~\cite{li2025eagle}, which improves earlier methods by incorporating tree-based attention and context-aware draft scoring. This allows the draft model to generate multiple candidate continuations per step while maintaining low latency, significantly reducing the number of serial decoding steps of the target model.

The draft model was trained on a carefully constructed dataset containing medical dialogue, clinical notes, and structured medical knowledge resources. To generate high-quality synthetic training data reflective of real-world medical interactions, we generated contextually relevant medical responses from Baichuan-M2, resulting in a diverse and domain-specific corpus. 

When deployed on a single RTX 4090 GPU with 4-bit quantization and a 4096-token prompt, the draft model achieved 73\% prediction accuracy and an average accepted length of 3.28 tokens per round. This resulted in a throughput increase from 41.5 to 89.9 tokens/s, a 2.17× speedup, demonstrating strong efficiency gains for text generation.

\section{Conclusion}

We have developed a dynamic reinforcement learning validation system that bridges the gap between the LLM evaluation and real-world clinical practice. This system replaces traditional static benchmarks with interactive patient simulations and multi-dimensional clinical assessment criteria, thereby creating a decision-making environment that closely mirrors real-world clinical scenarios. Using this innovative approach, we built and open-sourced the Baichuan-M2 model, which was trained using domain adaptation and multi-stage reinforcement learning. Despite having only 32 billion parameters, the Baichuan-M2 model demonstrates superior clinical reasoning capabilities. On the challenging HealthBench benchmark, the Baichuan-M2 model outperformed all other open-source models and rivaled leading closed-source systems, becoming one of only two models worldwide to achieve a score above 32 on the HealthBench Hard subset. Our work highlights that complex clinical performance can be achieved at a deployable scale, underscoring the potential of LLMs to significantly enhance clinical decision-making.

\section{Limitation and Future Work}

While we approach medical scenarios with deep reverence, we remain acutely aware that the journey toward using AI to improve human health is still a long and complex one. Despite our achievements, Baichuan-M2 is not without limitations that reflect the current state of technology. The model may still exhibit response hallucinations and insufficient reasoning stability in certain edge cases. From a metrics perspective, whether on HealthBench or other real-world medical capability evaluations, Baichuan-M2's performance is far from saturated, leaving considerable room for optimization across various clinical dimensions. Functionally, this version has not been fully optimized for capabilities such as tool calling and external knowledge retrieval, which could further enhance its clinical utility. We acknowledge these limitations transparently and commit to addressing them with a prudent and pragmatic approach, continuously refining the model's safety, reliability, and practical applicability in subsequent iterations.

Our current version primarily focuses on clinical diagnosis and treatment capabilities, but we recognize that medical inquiry skills and hallucination mitigation are equally critical for real-world deployment. Moving forward, we will strengthen quantitative assessment and optimization of these essential capabilities. Additionally, we plan to enhance research and implementation of multi-turn session reinforcement learning, aiming to provide comprehensive inquiry and diagnostic capabilities that mirror the complete clinical workflow. We also intend to explore advanced techniques for medical knowledge grounding, potentially integrating with medical knowledge bases and clinical decision support systems to further reduce hallucination rates and improve diagnostic accuracy.

\section{Contribution}

Contributors are presented in alphabetical order according to their first names. An asterisk (*) denotes those who are no longer part of the team.

\subsection*{Core Contributors}

Chengfeng Dou, 
Chong Liu*, 
Fan Yang, 
Fei Li, 
Jiyuan Jia, 
Mingyang Chen, 
Qiang Ju, 
Shuai Wang, 
Shunya Dang, 
Tianpeng Li, 
Xiangrong Zeng, 
Yijie Zhou

\subsection*{Contributors}

Chenzheng Zhu*, 
Da Pan, 
Fei Deng, 
Guangwei Ai, 
Guosheng Dong, 
Hongda Zhang, 
Jinyang Tai, 
Jixiang Hong*, 
Kai Lu, 
Linzhuang Sun, 
Peidong Guo, 
Qian Ma*, 
Rihui Xin, 
Shihui Yang, 
Shusen Zhang, 
Yichuan Mo, 
Zheng Liang

\subsection*{Experts and Advisors}

Xiaochuan Wang,
Zuyi Zhu, 
Hengfu Cui, 
Zhishou Zhang

\newpage
\bibliography{bibliography}
\bibliographystyle{unsrtnat}

\appendix

\section{Example of Evaluation Prompt for Rubrics}
\label{app:rubric-prompt}

Here are examples of evaluation prompts for positive and negative rubrics, where \texttt{<conversation>} denotes the conversation history between the user and the assistant, and \texttt{<rubric\_item>} denotes a rubric expressed in natural language with its corresponding criteria and point values.

\begin{tcolorbox}[title = {Example of Evaluation Prompt for Positive Rubrics}, breakable, fontupper=\ttfamily, fonttitle=\small]

Your task is to review a conversation along with a rubric item, and evaluate the final assistant response (i.e., the most recent turn) on how well it satisfies the rubric item.
The rubric item provided is positive, meaning it specifies what constitutes an acceptable response.
\\
\\
\# Conversation

<conversation>
\\
\\
\# Rubric item

<rubric\_item>
\\
\\
\# Instructions

Return a json object with the following fields: "explanation" and "acceptable".

- The "explanation" field should be a string explaining why the response does or does not meet the criterion and whether it is acceptable.

- The "acceptable" field should be a boolean indicating whether the response meets the criterion and is acceptable.
\\
\\
\# Response Format

Return just the json object in markdown format. Do not include any other text in the response.

\begin{verbatim}
{
  "explanation": "[You explanation here]",
  "acceptable": [true/false]
}
\end{verbatim}
\end{tcolorbox}

\begin{tcolorbox}[title = {Example of Evaluation Prompt for Negative Rubrics}, breakable, fontupper=\ttfamily, fonttitle=\small]

Your task is to review a conversation along with a rubric item, and evaluate the final assistant response (i.e., the most recent turn) on how well it satisfies the rubric item.
The rubric item provided is negative, meaning it specifies what constitutes an unacceptable response.
\\
\\
\# Conversation

<conversation>
\\
\\
\# Rubric item

<rubric\_item>
\\
\\
\# Instructions

Return a json object with the following fields: "explanation" and "unacceptable".

- The "explanation" field should be a string explaining why the response does or does not meet the criterion and whether it is unacceptable.

- The "unacceptable" field should be a boolean indicating whether the response meets the criterion and is unacceptable.
\\
\\
\# Response Format

Return just the json object in markdown format. Do not include any other text in the response.

\begin{verbatim}
{
  "explanation": "[You explanation here]",
  "unacceptable": [true/false]
}
\end{verbatim}
\end{tcolorbox}

\section{Response Case of HealthBench}
\label{app:case-healthbench}

\begin{figure*}[t]
\centering
\includegraphics[width=1.0\textwidth]{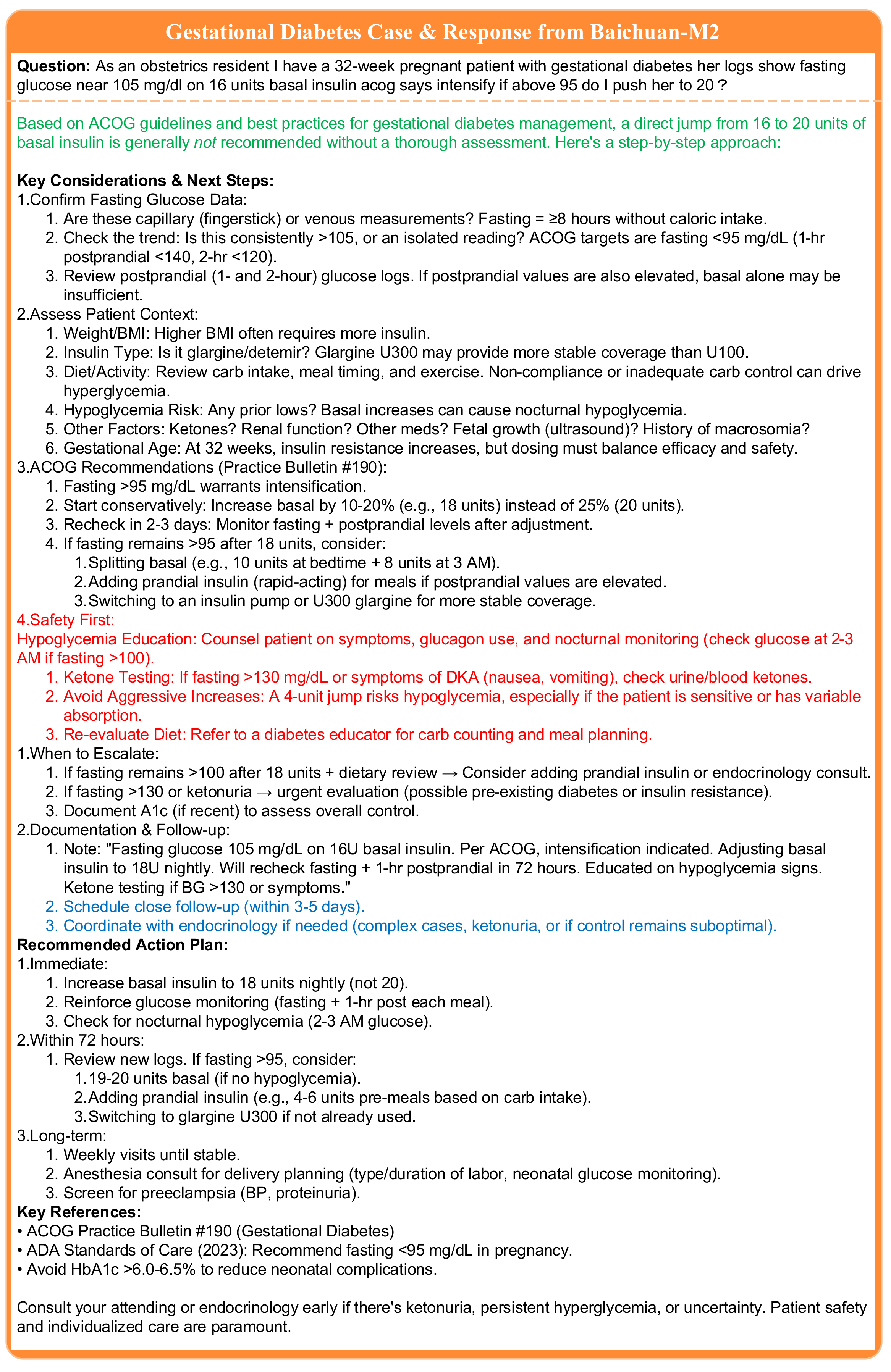}
\caption{Case of Gestational Diabetes answered by Baichuan-M2, which shows superior performance in \textcolor{ForestGreen}{Accuracy}, \textcolor{Maroon}{Relevance}, and \textcolor{RoyalBlue}{Completeness}.}
\label{fig:case1_m2}
\end{figure*}

\begin{figure*}[t]
\centering
\includegraphics[width=1.0\textwidth]{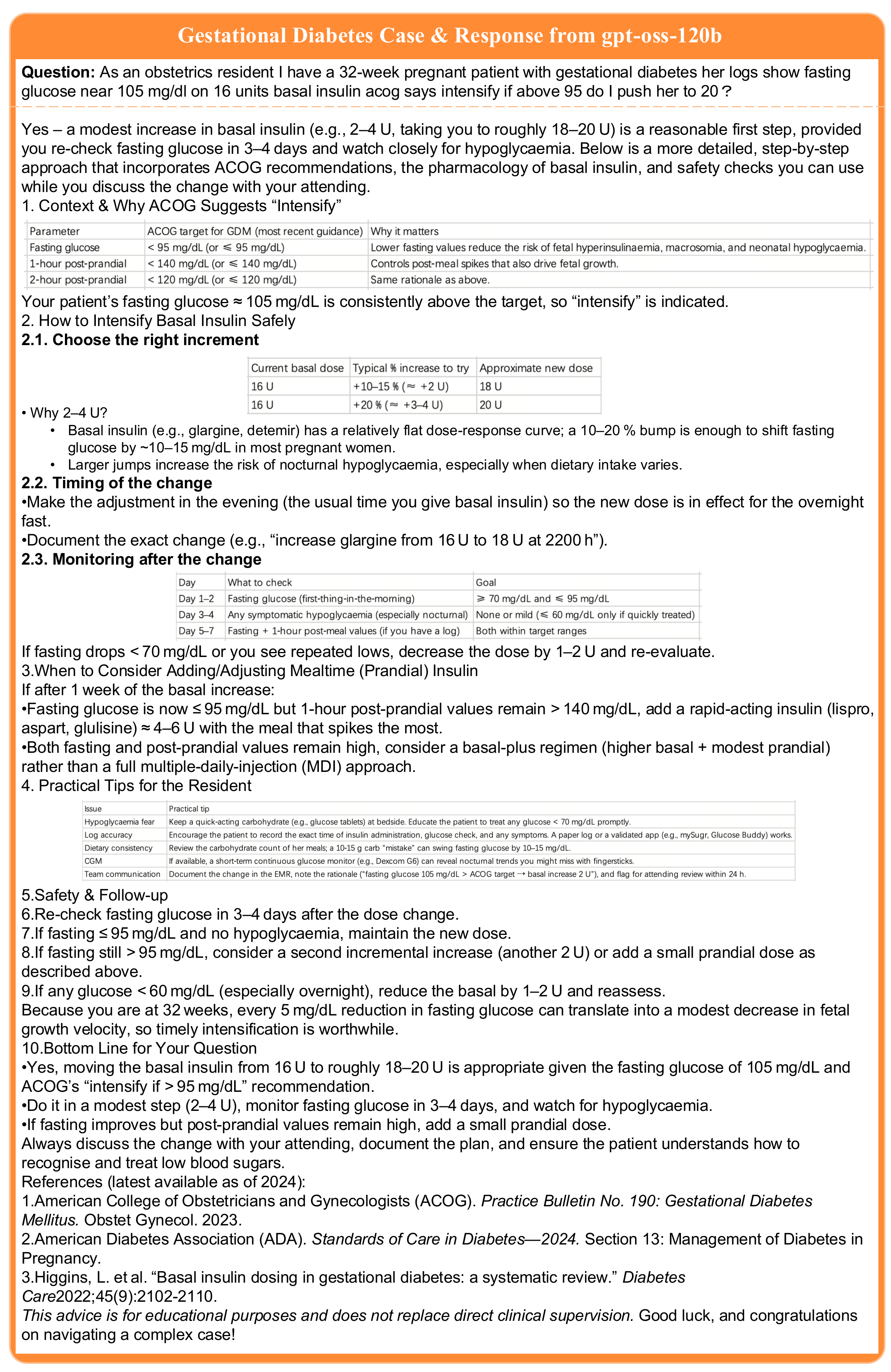}
\caption{Gestational Diabetes case responded by gpt-oss-120b.}
\label{fig:case1_gpt}
\end{figure*}

\end{document}